\documentclass[11pt]{article}
\usepackage{acl2016}
\usepackage{times}
\usepackage{url}
\usepackage{latexsym}
\usepackage{graphicx}
\usepackage[farskip=1px]{subfig}
\usepackage[inline]{enumitem}
\setlist[enumerate]{itemsep=0px, parsep=0px}
\setlist[description]{itemsep=2px, parsep=0px, leftmargin=0cm, topsep=2px}
\usepackage{booktabs}
\usepackage{expex}
\lingset{labeloffset=1ex,textoffset=3ex}
\lingset{aboveexskip=5px,belowexskip=5px,interpartskip=0px}
\usepackage{todonotes}
\usepackage{verbatim}

\usepackage[hidelinks]{hyperref}
\usepackage{url}

\PassOptionsToPackage{hyphens}{url}
\expandafter\def\expandafter\UrlBreaks\expandafter{\UrlBreaks
  \do\a\do\b\do\c\do\d\do\e\do\f\do\g\do\h\do\i\do\j%
  \do\k\do\l\do\m\do\n\do\o\do\p\do\q\do\r\do\s\do\t%
  \do\u\do\v\do\w\do\x\do\y\do\z\do\A\do\B\do\C\do\D%
  \do\E\do\F\do\G\do\H\do\I\do\J\do\K\do\L\do\M\do\N%
  \do\O\do\P\do\Q\do\R\do\S\do\T\do\U\do\V\do\W\do\X%
  \do\Y\do\Z\do\*\do\-\do\~\do\'\do\"\do\-}%

\aclfinalcopy

\title{Pragmatic factors in image description: the case of negations}

\author{Emiel van Miltenburg\\
Vrije Universiteit Amsterdam\\
{\tt emiel.van.miltenburg@vu.nl} 
\And
Roser Morante\\
Vrije Universiteit Amsterdam\\
{\tt roser.morante@vu.nl}
\AND
Desmond Elliott\\
ILLC, University of Amsterdam\\
{\tt d.elliott@uva.nl}
}

\date{}

\begin{document}

\maketitle

\begin{abstract}
We provide a qualitative analysis of the descriptions containing negations (\emph{no}, \emph{not}, \emph{n't}, \emph{nobody}, etc) in the Flickr30K corpus, and a categorization of negation uses. Based on this analysis, we provide a set of requirements that an image description system should have in order to generate negation sentences. As a pilot experiment, we used our categorization to manually annotate sentences containing negations in the Flickr30k corpus, with an agreement score of $\kappa$=0.67. With this paper, we hope to open up a broader discussion of subjective language in image descriptions.

\end{abstract}

\section{Introduction}

Descriptions of images are typically collected from untrained workers via crowdsourcing platforms, such as Mechanical Turk\footnote{\url{http://www.mturk.com}}. 
The workers are explicitly instructed to describe only what they can see in the image, in an attempt to control content selection \cite{young2014image,chen2015}. 
However, workers are still free to project their world view when writing the descriptions and they make linguistic choices, such as using negation structures \cite{van2016stereotyping}.

In this paper we study the use of \emph{negations} in image descriptions. 
A negation is a word that communicates that something is \emph{not} the case. 
Negations are often used when there is a mismatch between what speakers 
expect to be the case and what is actually the case (see e.g. \cite{leech1983principles,beukeboom2010negation}). For example, if Queen Elizabeth of England were to appear in public wearing jeans instead of a dress, (\nextx a) would be acceptable because she is known to wear dresses in public. But if she were to show up wearing a dress, (\nextx b) would be unexpected. 

\pex
\a Queen Elizabeth isn't wearing a dress
\a \ljudge{??} Queen Elizabeth isn't wearing jeans
\xe
\lingset{labeloffset=1ex,textoffset=1ex}

Thus the correct use of negations often requires \emph{background knowledge}, or at least some sense of what is expected and what is not.

We focus on two kinds of negations: \textbf{non-affixal negations} (\emph{not, n't, never, no, none, nothing, nobody, nowhere, nor, neither}) \cite{tottie1980affixal}; and \textbf{implicit negations} in the form of prepositions (\emph{without, sans,} and \emph{minus}), and the verbs \emph{lack, omit, miss} and \emph{fail}. \newcite{horn1979negation} calls this second category `inherent negatives'. 
Affixal negations (words starting with \emph{a--, dis--, un--, non--, un--} or ending with \emph{--less}) are beyond the scope of this paper, but we hope to address them in future work. 

The main contributions of this paper are an overview of different uses of negations in image description corpora, analysing the background knowledge required to generate negations, and the implications for image description models.\footnote{We provide all of our code, data, and annotation guidelines online. See: \url{https://github.com/evanmiltenburg/annotating-negations}}
\section{Data}
We focus on negations on the Flickr30K dataset \cite{young2014image}.
The negations were detected by lexical string-matching using regular expressions, except for the verbs. 
For the verbs, we checked if any of the tokens starts with \emph{lack, omit, miss} or \emph{fail}. Our search yielded 896 sentences, of which 892 unique, and 31 false positives. Table \ref{table:negfrequency} shows frequency counts for each  negation term.

\begin{table}[htb]
\centering
\begin{tabular}{p{1.1cm}r p{1.2cm}r p{1.3cm}r}
\toprule
 no      & 371 &  nothing &  16 &  neither &   2 \\
 not     & 198 &  lack    &   9 &  sans    &   1 \\
 without & 141 &  fail    &   9 &  none    &   1 \\
 miss    &  69 &  never   &   5 &  nobody  &   1 \\
 n't     &  68 &  nowhere &   3 & & \\
\bottomrule
\end{tabular}
\caption{Frequency counts for each  negation term. }\label{table:negfrequency} 
\end{table}

We carried out the same analysis for the Microsoft COCO dataset \cite{chen2015} to see if the proportion of negations is a constant. Our approach yielded yielded 3339 sentences on the training and validation splits, of which 3232 unique. The presence of negations appears to be a linear function of dataset size: 0.56\% in the Flickr30K dataset, and 0.54\% in the MS COCO dataset. This suggests that the use of negations is not particular to either dataset, but rather it is a robust phenomenon across datasets.


Table \ref{table:imageswithnegations} shows the distribution of descriptions containing negations across images.
 In the majority of cases only one of the five descriptions contains a negation (86.25\% in Flickr30K and 72.05\% in MS COCO). Only in very exceptional cases do the five descriptions contain negations. This indicates that the use of negation is a subjective choice.

\begin{table}[h!]
\centering
\begin{tabular}{lrrrrr}
\toprule
\textbf{Dataset} & \textbf{1} & \textbf{2} & \textbf{3}& \textbf{4} & \textbf{5}\\\midrule
Flickr30K & 659& 85& 16& 1& 3\\
MS COCO & 2406& 277& 78& 30& 5\\
\bottomrule
\end{tabular}
\caption{Distribution of the number of descriptions of an image with at least one negation term.}\label{table:imageswithnegations}
\end{table}
\section{Negation uses in image descriptions}\label{sec:negations}

In this section, we provide a categorization of negation uses and assess the amount of required background knowledge for each use. Our categorization is the result of manually inspecting all the data twice: the first time to develop a taxonomy, and the second time to apply this taxonomy to all 892 sentences. Note that our categorization is meant as a \emph{practical guide} to be of use for natural language generation. There is already a unifying explanation for \emph{why} people use negations (unexpectedness, see \cite{leech1983principles,beukeboom2010negation}). The question here is \emph{how} people use negations, what they negate, and what kind of knowledge is required to produce those negations. 


\noindent\textbf{Salient absence}: 
The first use of negation is to indicate that something is absent:

\pex
\a A man \textbf{without} a shirt playing tennis. \label{ex:shirtless}
\a A woman at graduation \textbf{without} a cap on.\label{ex:capless}
\xe
\vspace{-0.4cm}

Shirts and shoes are most commonly mentioned as being absent in the Flickr30K dataset. From examples like (\lastx a) speaks the norm that people are supposed to be fully dressed. These examples seem common enough for a machine to learn the association between exposed chests and the phrase \emph{without a shirt}. But there are also more difficult cases, such as (\lastx b). To describe an image like this, one should know that students (in the USA) typically wear caps at their graduation. This example shows the importance of background knowledge for the full description of an image.\\

\begin{figure}[h]
\centering
\includegraphics[width=0.3\textwidth]{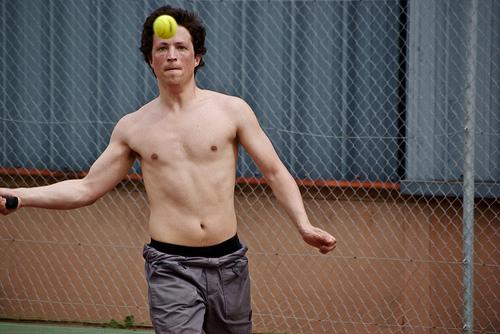}\\
Example \ref{ex:shirtless} (Image 2883099128)
\end{figure}

\noindent\textbf{Negation of action/behavior}: 
The second category is the use of negation to deny that an action or some kind of behavior is occurring:

\pex
\a A kid eating out of a plate \textbf{without} using his hands.\label{ex:withouthands}
\a A woman in the picture has fallen down and \textbf{no} one is stopping to help her up.
\xe
\vspace{-0.4cm}

Examples like these require an understanding of what is likely or supposed to happen, or how people are expected to behave.\\

\begin{figure}[h]
\centering
\includegraphics[width=0.3\textwidth]{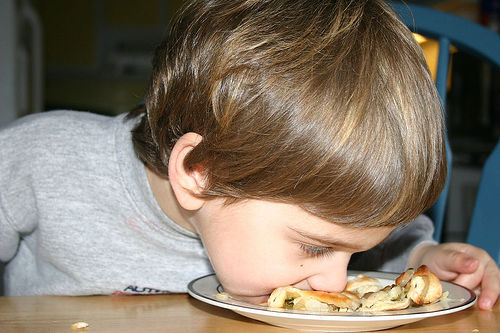}\\
Example \ref{ex:withouthands} (Image 39397486)
\end{figure}

\noindent\textbf{Negation of property}: 
The next use of negation is to note that an entity in the image lacks a property. In (\nextx a), the negation does two things: it highlights that the buildings are not finished, but in its combination with \emph{yet} suggests that they \emph{will be} finished.

\pex
\a A man wearing a hard hat stands in front of buildings \textbf{not} yet finished being built.\label{ex:hardhat}
\a There are four boys playing soccer, but \textbf{not} all of them are on the same team [\ldots].
\xe
\vspace{-0.4cm}

In (\lastx b), the negated phrase also performs two roles: it communicates that there are (at least) two teams, and it denies that the four boys are all in the same team. For both examples, the negated parts (\emph{being finished} and \emph{being on the same team}) are properties associated with the concepts of \textsc{building} and \textsc{playing together}, and could reasonably be expected to be true of buildings and groups of boys playing soccer. The negations ensure that these expectations are cancelled.




\begin{figure}[h]
\centering
\includegraphics[width=0.3\textwidth]{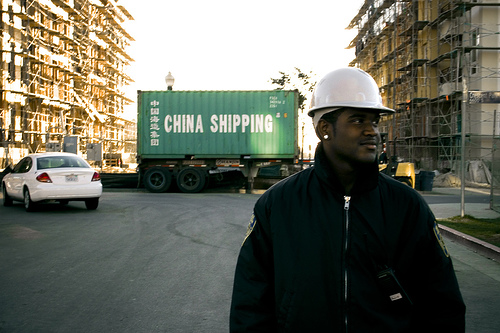}\\
Example \ref{ex:hardhat} (Image 261883591)
\end{figure}

Example (\nextx) shows a completely different effect of negating a property. Here, the negation is used to \emph{compare} the depicted situation with a particular \emph{reference point}. The implication here is that the picture is not taken in the USA.

\ex
A wild animal \textbf{not} found in america jumping through a field.
\xe

\noindent\textbf{Negation of attitude}:
The fourth use of negation concerns attitudes of entities toward actions or others. The examples in (\nextx) illustrate that this use requires an understanding of emotions or attitudes, but also some reasoning about what those emotions are directed at.

\pex
\a A man sitting on a panel \textbf{not} enjoying the speech.\label{ex:panel}
\a The dog in the picture does\textbf{n't} like blowing dryer.
\xe

\begin{figure}[h]
\centering
\includegraphics[width=0.3\textwidth]{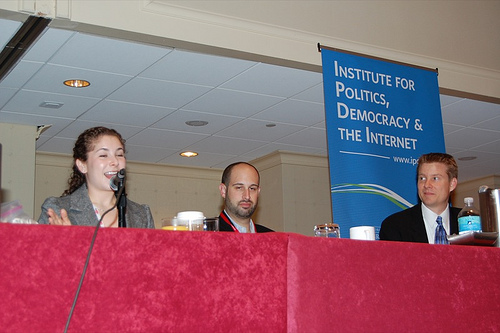}\\
Example \ref{ex:panel} (Image 2313609814)
\end{figure}

\noindent\textbf{Outside the frame}:
The most image-specific use of negation is to note that particular entities are not depicted or out of focus:

\pex
\a A woman is taking a picture of something \textbf{not} in the shot with her phone.\label{ex:takingpicture}
\a Several people sitting in front of a building taking pictures of a landmark \textbf{not} seen.
\xe
\vspace{-0.4cm}

The use of negation in this category requires an understanding of the events taking place in the image, and what entities might be involved in such events. (\lastx b) is a particularly interesting case, where the annotator specifically says that there is a \emph{landmark} outside the frame. This raises the question: how does she know and how could a computer algorithm recognise this?

\begin{figure}[h]
\centering
\includegraphics[width=0.3\textwidth]{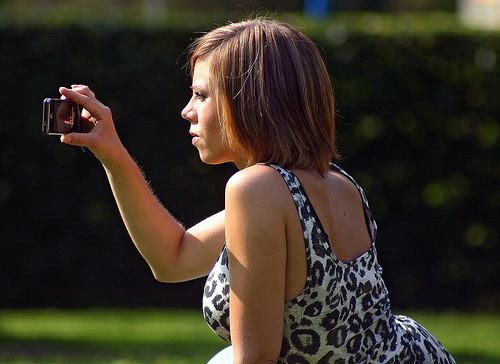}\\
Example \ref{ex:takingpicture} (Image 4895028664)
\end{figure}

\noindent\textbf{(Preventing) future events}: 
The sixth use of negation concerns future events, generally with people preventing something from happening. Here are two examples:

\pex
\a A man is riding a bucking horse trying to hold on and \textbf{not} get thrown off.\label{ex:horse}
\a A girl tries holding onto a vine so she wo\textbf{n't} fall into the water.
\xe
\vspace{-0.4cm}

What is interesting about these sentences is that the ability to produce them does not only require an understanding of the depicted situation (someone is holding on to a horse/vine), but also of the possibilities within that situation (they may or may not fall off/into the water), depending on the actions taken.

\begin{figure}[h]
\centering
\includegraphics[width=0.3\textwidth]{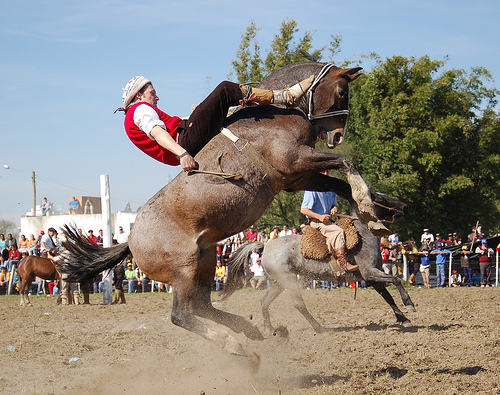}\\
Example \ref{ex:horse} (Image 263428541)
\end{figure}

\noindent\textbf{Quotes and Idioms}: 
Some instances of negations are \emph{mentions} rather than \emph{uses} as shown in (\nextx).

\ex
A girl with a tattoo on her wrist that reads ``\textbf{no} regrets'' has her hand outstretched.
\xe
\vspace{-0.4cm}

Other times, the use of a negation isn't concerned with the image as much as it is with the English language. The examples in (\nextx) illustrate this \emph{idiomatic} or \emph{conventional} use of negation. 

\pex
\a Strolling down path to \textbf{nowhere}.\label{ex:pathtonowhere}
\a Three young boys are engaged in a game of do\textbf{n't} drop the melon.
\xe

\begin{figure}[h]
\centering
\includegraphics[width=0.3\textwidth]{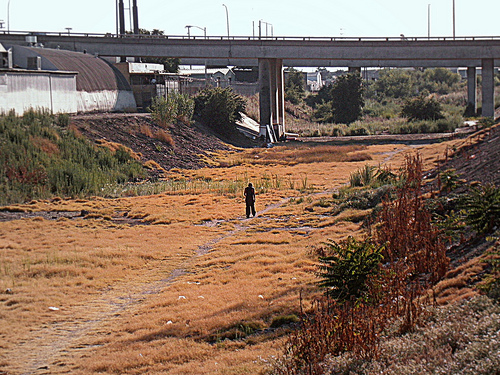}\\
Example \ref{ex:pathtonowhere} (Image 4870785283)
\end{figure}

\noindent\textbf{Other}: 
Several sentences do not fit in any of the above categories, but there aren't enough similar examples to merit a category of their own. Two examples are given below. In (\nextx), the negation is used to convey that it is \emph{atypical} to be holding an umbrella when it is not raining.

\lingset{textoffset=1ex}
\ex
The little boy [\ldots] is smiling under the blue umbrella even though it is \textbf{not} raining.\label{ex:littleboy}
\xe

\begin{figure}[h]
\centering
\includegraphics[width=0.3\textwidth]{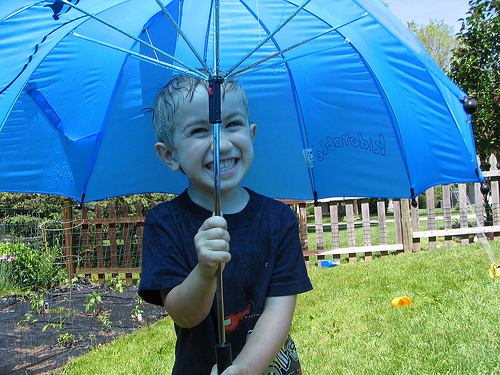}\\
Example \ref{ex:littleboy} (Image 371522748)
\end{figure}

In (\nextx), the annotator recognized the intention of the toddler, and is using the negation to contrast the goals with the ability of the toddler. Though there are many other sentences where the negation is used to contrast two parts of the sentence (see Section \ref{annotating}), there is just one example where an \emph{ability} is negated.

\ex
A little toddler trying to look through a scope but ca\textbf{n't} reach it.
\xe

We expect have no doubt that there are still other kinds of examples in the Flickr30K and the MS COCO datasets. Future research should assess the degree to which the current taxonomy is sufficient to systematically study the production of negations in image descriptions.

\section{Annotating the Flickr30K corpus}\label{annotating}
Two of the authors annotated the Flickr30K corpus using the categories listed above with two goals: to validate the categories, and to develop annotation guidelines for future work. By going through all sentences with negations, we were able to identify borderline cases that could serve as examples in the final guidelines.

Using the categories defined in Section \ref{sec:negations}, we achieved an inter-annotator agreement of Cohen's $\kappa$=0.67, with an agreement of 77\%. We then looked at sentences with disagreement, and settled on categories for those sentences. Table \ref{table:frequencies} shows the final counts for each category, including a Meta-category for cases like \emph{I don't see a picture}, commenting on the original annotation task, or on the images without describing them. 

\begin{table}[htb]
\centering
\begin{tabular}{lr}
\toprule
\textbf{Category} & \textbf{Count}\\
\midrule
 Salient absence             & 488 \\
 Negation of action/behavior &  90 \\
 Quotes and idioms           &  71 \\
 Not a description/Meta      &  40 \\
 Negation of attitude        &  36 \\
 False positive              &  31 \\
 Outside the frame           &  26 \\
 Negation of property        &  25 \\
 (Preventing) future events  &  21 \\
 Other                       &  66 \\
\bottomrule
\end{tabular}
\caption{Frequency count of each category.}\label{table:frequencies}
\end{table}

In addition to our categorization, we found 39 examples where negations are also used to provide \textbf{contrast} (next to their use in terms of the categories listed above). Two examples are:

\pex
\a A man shaves his neck but \textbf{not} his beard
\a A man in a penguin suit runs with a man, \textbf{not} in a penguin suit
\xe
\vspace{-0.4cm}

Such examples show how negations can be used to structure an image. Sometimes this leads to a scalar implicature \cite{horn1927semantic}, like in (\ref{ex:implicature}).

\ex \label{ex:implicature}
Three teenagers, two \textbf{without} shoes having a water gun fight with various types of guns trying to spray each other.\\ $\Rightarrow$ One teenager \emph{is} wearing shoes.
\xe
\vspace{-0.4cm}

A striking observation is that many negations pertain to pieces of clothing; for example: 282 (32\%) of the negations are about people being shirtless, while 59 (7\%) are about people not wearing shoes. It is unclear whether this is due to selection bias, or whether the world just contains many shirtless people. But we expect that this distribution will make it difficult for systems to learn how to use negations that aren't clothing-related.
\section{Discussion}


The negations used by crowdworkers are likely to have required some form of ``world knowledge''. We now discuss potential sources of 
evidence for recognising a candidate for negation in the description of an image:
\begin{enumerate*}[label=(\alph*)]
  \item The \emph{Outside the frame} category requires an understanding of human gaze within an image, which is a challenging problem in computer vision \cite{valenti2012}. Additionally, we also need to understand the differences between scene types, both from a computational- \cite{oliva2001} and a human perspective \cite{torralba2006}. 
  \item The \emph{Salient absence} category provides evidence for two kinds of expectations that play a role in the use of negations: general expectations (people are supposed to wear shirts, cf.\ \ref{ex:shirtless}) and situation-specific expectations (students at graduation ceremonies typically wear caps, cf.\ \ref{ex:capless}).
  \item Finally, the \emph{Negation of action/behavior} category requires action recognition, which is a challenging problem in still images \cite{poppe2010}. The ability to automatically recognise what people are doing in an image, and how this contrasts with what they would typically do in similar images, would greatly help with generating this use of negation.
\end{enumerate*}

From a linguistic perspective, background knowledge could be represented by \emph{frames} \cite{fillmore1976frame} and \emph{scripts} \cite{schank1977scripts}. There are some hand-crafted resources that contain this kind of knowledge, e.g. FrameNet \cite{baker1998berkeley}, but they only have limited coverage. Recent work has shown, however, that it is possible to automatically learn frames \cite{pennacchiotti2008automatic} and script knowledge  \cite{Chambers:2009:ULN:1690219.1690231} from text corpora. \newcite{fast2016augur} show how such knowledge, as well as knowledge about \emph{object affordances} \cite{gibson1977theory}, can be used to reason about visual scenes.
\section{Conclusion}

We studied the use of negations in the Flickr30K dataset. The use of negations imply that the descriptions contain a combination of objective and subjective interpretations of the images. But negations are only one type of subjective language in image description datasets. We expect that different subjective language use (e.g.\ discourse markers such as \emph{yet} or \emph{even though}) can be observed with relative ease in this and other datasets. Additionally it would be interesting to study the use of negations in different languages, such as the German-English Multi30K dataset \cite{2016arXiv160500459E}. We encourage further research to discover other types of subjective language in vision and language datasets, and studies of how subjective language may affect language generation.
\section{Acknowledgments}
EM and RM are supported by the Netherlands Organization for Scientific Research (NWO) via the Spinoza-prize awarded to Piek Vossen (SPI 30-673, 2014-2019).  DE is supported by NWO Vici grant nr. 277-89-002 awarded to Khalil Sima'an. 


\bibliographystyle{acl2016}
\bibliography{negation_refs}
\end{document}